# An Integrated Fusion Framework for Ensemble Learning Leveraging Gradient Boosting and Fuzzy Rule-Based Models

Jinbo Li, Peng Liu, Long Chen, *Senior Member, IEEE*, Witold Pedrycz, *Life Fellow, IEEE*, Weiping Ding, *Senior Member, IEEE*

*Abstract*—The integration of different learning paradigms has long been a focus of machine learning research, aimed at overcoming the inherent limitations of individual methods. Fuzzy rule-based models excel in interpretability and have seen widespread application across diverse fields. However, they face challenges such as complex design specifications and scalability issues with large datasets. The fusion of different techniques and strategies, particularly Gradient Boosting, with Fuzzy Rule-Based Models offers a robust solution to these challenges. This paper proposes an Integrated Fusion Framework that merges the strengths of both paradigms to enhance model performance and interpretability. At each iteration, a Fuzzy Rule-Based Model is constructed and controlled by a dynamic factor to optimize its contribution to the overall ensemble. This control factor serves multiple purposes: it prevents model dominance, encourages diversity, acts as a regularization parameter, and provides a mechanism for dynamic tuning based on model performance, thus mitigating the risk of overfitting. Additionally, the framework incorporates a sample-based correction mechanism that allows for adaptive adjustments based on feedback from a validation set. Experimental results substantiate the efficacy of the presented gradient boosting framework for fuzzy rule-based models, demonstrating performance enhancement, especially in terms of mitigating overfitting and complexity typically associated with many rules. By leveraging an optimal factor to govern the contribution of each model, the framework improves performance, maintains interpretability, and simplifies the maintenance and update of the models.

*Impact Statement*—In today's world of fast-paced advancements in artificial intelligence, understanding and improving the effectiveness of models is incredibly important. Our research introduces a new way of combining models based on fuzzy rules with a technique called gradient boosting, aiming to balance the clarity of the model with its performance. This balance is particularly crucial in fields like healthcare, finance, and policy-making, where making accurate and understandable decisions is essential. Our approach includes a dynamic control factor, which not only prevents any single model from becoming too dominant but also encourages diversity, serving as a sort of adjustment knob. Moreover, our model can adjust itself dynamically based on feedback from a set of validation data, a feature not typically found in traditional systems. Experimental results show that our method is effective at reducing overfitting and complexity, two common challenges in AI models.

## I. INTRODUCTION

UNDERSTANDING the sophisticated relationships within datasets represents a fundamental challenge and a critical research avenue. Data mining, with its array of methodologies such as association rule mining, cluster analysis, and statistical analysis, serves as a cornerstone for analyzing and interpreting complex data structures [1], [2], [3], [4], [5]. For instance, association rule mining methods [6], [7] excel in identifying frequent item sets, clustering algorithms [8], [9] are adept at recognizing sample characteristics and mining similarities across categories, hierarchical analysis[10], [11] offers deep insights into complex datasets by breaking them down into smaller, more manageable components, and statistical analysis employs a range of techniques like descriptive statistics, probability theory, and inferential statistics to draw meaningful conclusions from aggregated data. Recent advancements in technology, particularly in deep learning and its variants [1], [5], [12], have demonstrated significant success in addressing regression problems. However, these advancements also introduce a complexity in understanding the models' internal mechanisms and the rationale behind their predictions [4], [13]. This complexity underscores the necessity for data mining methods that not only harness computational power but also ensure the interpretability and accessibility of the insights they generate. Therefore, this study is motivated by the objectives of leveraging the analytical capabilities of data mining techniques and enhancing the clarity and utility of the insights derived from them.

The Takagi-Sugeno (TS) inference system [14], [15] is a

Jinbo Li is with the Research Institute, China Unicom, Beijing 100048, China (e-mail: lijb7213@chinaunicom.cn).

Peng Liu is with the International Business School, Henan University, China (e-mail: liupeng@henu.edu.cn)

Long Chen is with the Department of Computer and Information Science, University of Macau, Macau 999078, China. (e-mail: longchen@umac.mo)

Witold Pedrycz is with the Department of Electrical and Computer Engineering, University of Alberta, Edmonton, AB T6R 2V4, Canada (e-mail: wpedrycz@ualberta.ca)

Weiping Ding is with the School of Information Science and Technology, Nantong University, Nantong 226019, China (e-mail: dwp9988@163.com).

* (Corresponding author: Weiping Ding).

fuzzy logic system that makes choices employing if-then rules to breakdown a complex problem into a succession of relatively simple ones. In general, each rule has an antecedent and a consequent part, with the antecedent component explicitly specifying what conditions must be met before an action may be performed and the consequent part outlining precisely what the action would require [16], [17], [18]. When compared to other models, the IF-Then rule has a better degree of interpretability since it is compatible with the typical human decision-making process. In the financial domain, fuzzy rule-based systems leverage a set of fuzzy rules to assess the risk level of loan applicants with greater interpretability. For instance, an applicant with a good credit record but unstable income might be assigned a medium risk level. In contrast, applicants with 'good credit records and stable income' and 'poor credit records and unstable income' could be evaluated as having high and low risk levels, respectively. Similarly, in the healthcare industry, symptoms, vital signs, and laboratory results can be integrated into a series of fuzzy rules to assist physicians in making preliminary diagnoses. For example, frequent coughing and high fever could indicate influenza. Thus, fuzzy rule-based systems offer understandable decision support across various domains by aligning closely with human reasoning processes. Although TS models have shown impressive performance in a variety of domains, TS systems struggle to simulate complicated systems using a limited set of rules, particularly when dealing with high-dimensional data. On the other hand, when there are an excessively considerable number of If-Then rules, the model's interpretability in a particular domain becomes more challenging [19], [20], [21].

Gradient boosting, as an integrated learning strategy that combines various weak learners for learning, can increase the model's accuracy and efficacy as well as its generalization capabilities [22], [23], [24]. In particular, instead of starting from scratch, weak learners might learn from the outcomes produced by learners in the preceding phase of the model learning process. The current weak learner learns the previous learner's output to conduct model correction, minimizing variance to improve model accuracy and eventually achieving the strong model [23], [24]. Gradient enhancement has been effectively used in a couple of areas; for instance, gradient enhancement may be used to detect and categorize consumer behavior patterns. This enables companies to better comprehend their clients and customize their offerings to their requirements. Gradient boosting is also increasingly utilized in image processing, natural language processing, robotics, the Internet of Things, and other domains [23], [24], [25].

The primary objective of this research is the development of a Fusion Framework that integrates Gradient Boosting methods with Fuzzy Rule-Based Models, positioning the latter as the base learners. This initiative is propelled by a trio of motivations: the interpretability of fuzzy rule-based models ensures their versatility across various domains; their proficiency in capturing local data patterns through individual fuzzy rules representing specific input space segments; and their modular nature, which supports the assembly of complex systems from simpler elements, enhancing scalability by allowing the addition of rules in response to new data or insights.

One core of our framework is the implementation of a dynamically-adjustable control factor during each iteration of model formulation. This factor serves multiple critical functions: it prevents any model from overwhelming the final decision, thus averting overfitting; it introduces diversity by assigning distinct control factors to each model, improving the ensemble's generalization ability; it acts as a regularization mechanism, limiting model complexity to further resist overfitting; and it allows for adjustments based on model performance, offering a safeguard against initial phase overfitting due to data imbalances or noise. Additionally, the Fusion Framework employs a sample-based correction strategy, utilizing feedback from a validation set to dynamically refine model complexity. This includes simplifying models if their performance on the validation set is lacking, addressing the risk of overtraining. The framework also uses the validation set to optimize the contribution of each model, reducing the weight of models that underperform on the challenge set, thus indicating potential overfitting to the training data. This structured approach ensures that our Fusion Framework not only leverages the strengths of fuzzy rule-based models but also maintains a balance between complexity and interpretability, fostering robustness against overfitting.

The originality and contribution of this research work primarily emanates from the following areas:

1) The proposed framework uniquely fuses multiple fuzzy rule-based models by using gradient boosting algorithm, crafting a powerful ensemble well-suited for regression challenges. It leverages interpretability and local pattern recognition, with each fuzzy rule encapsulating knowledge about specific regions of the input space.

2) Building upon the modular nature of fuzzy rules, the framework exhibits a fusion-centric scalable design. This allows for the incremental addition of new models or rules, adapting to varying complexities and sizes of datasets. Essentially, the fusion of these elements brings an important level of adaptability and scalability, an innovation in the design of fuzzy rule-based systems.

3) One of the original contributions lies in the dynamic control factor that modulates each model's impact within the ensemble. This control factor performs multiple roles: it mitigates the risk of overfitting, fosters diversity among base learners, and augments the generalization performance of the overall system. Moreover, by incorporating a sample-based correction mechanism that recalibrates model complexity according to validation set performance, the framework gains an additional layer of robustness and adaptability.

The structure of this article is organized as follows: In Section II, we offer an in-depth literature review of the most recent developments in fuzzy rule-based models and explore the incorporation of the concept of fuzzy sets into the domains including ensemble learning, incremental learning, and gradient-related algorithms. In Section III, we describe the architecture of our proposed model and delve into the intricate details of key components. In Section IV, we present our experimental results and finally draw the conclusion of the article.

## II. RELATED WORK

While the basic paradigm of fuzzy modeling is largely sound and valid, it can still be questioned based on the principle of incrementality in model development. This principle highlights the importance of progressively approaching the goal during model construction and the need to gradually optimize the model structure during model development. Following the principle of incrementality, all models should be constructed starting from the simplest and most general form to ensure that the basic framework of the model is sufficiently robust and adaptable. When needed, the models will be optimized and refined in subsequent stages by iterative means to better adapt to the constant changes in the real world. This stepwise optimization process helps to improve the prediction accuracy and adaptability of the model, making it more practical.

In the realm of regression problems, the utilization of fuzzy rule-based models has become a pervasive method of approach. Over the years, a multitude of researchers have put forward various types of fuzzy rule-based models, each tailored to serve the purpose at hand and tailored for optimal results [12], [13], [25], [26], [27]. These diverse models exhibit a broad spectrum of characteristics, and each one incorporates unique aspects that make it distinctive. Alongside the creation of these models, the development of corresponding parameter optimization techniques has been also a central focus.

*Advanced Optimization Strategies*: For instance, in [9], the FCM-RDpA, which blends fuzzy clustering, regularization, and Powerball Adabelief optimization techniques, is shifted towards independent membership function (MF) types of rule bases, as opposed to the formerly used shared membership functions, enhancing the system's aptitude for high-dimensional problem handling. The study [28] proposed a fuzzy time series (FTS) forecasting method that optimizes fuzzification, defuzzification, and incorporates an innovative error learning process. The method employs the FS criterion for optimal interval determination and defuzzification optimization. These strategies not only underscore the dynamic nature of optimization in fuzzy systems but also highlight the continuous effort to enhance system performance, interpretability, and applicability across various complex problem domains.

*Enhancing Interpretability in Fuzzy Systems*: The authors [16] introduced FIMG-TSK, a fully interpretable first-order TSK fuzzy system, integrating aspects of Gaussian mixture model (GMM). It tackles intricate training issues of TSK fuzzy systems, improving generalization via rule weight determination. The study [33] introduces a unique disjunctive fuzzy neural network approach to boost the interpretability of T-S fuzzy models for high-dimensional regression problems. An OR neural layer and a unique learning algorithm are incorporated to discern unknown parameters. these contributions represent an advance forward in the quest to improve the interpretability of fuzzy systems, addressing both the technical intricacies of model training and the overarching need for transparent, understandable machine learning solutions.

*Feature Engineering and Model Simplification*: In [29], A fuzzy inference system (FIS) methodology integrates Fuzzy C-Means (FCM) clustering and feature selection, offering an innovative approach to multiple linear regression models that can manage a considerable number of independent variables even with limited datasets. In [32], the authors aim to design an advanced fuzzy model for high-dimensional issues, focusing on reducing fuzzy rules and managing overfitting. It introduces a strategy using statistical analysis for input variable selection, allowing input space reduction while preserving the original structure. This assists in constructing two types of fuzzy models and formulating fuzzy rule's premises via fuzzy clustering, mitigating overfitting and rule reduction challenges. Together, these studies underscore the pivotal role of feature engineering and model simplification in advancing the practical applicability and theoretical understanding of fuzzy systems, particularly in the face of high-dimensional and sparse data scenarios.

*Evolution and Diversity of Fuzzy rule-based models*: The authors [8] develop a new method, incremental fuzzy clustering-based neural networks (IFCNNs), adept at handling incremental learning tasks and dynamic input space division. They propose a dynamic FCMs clustering algorithm for real-time updates with new data, enabling improved management of incremental data and robust generalization. The research [31] focuses on building fuzzy rule-based models within horizontal federated learning, considering data privacy. It presents a bifurcated federated learning strategy for training a global model on scattered data, eliminating data centralization and reducing user privacy and communication overheads. This trajectory of innovation reflects a deepened understanding of the complexities inherent in modern data environments and represents a critical stride towards the refinement of fuzzy systems in addressing such challenges.

Recent research trends have witnessed the ingenious incorporation of the concept of fuzzy sets into the domain of Ensemble Learning and Incremental Learning, especially within gradient-related algorithms (such as Gradient Descent, Gradient Boosting, Mini-Batch Gradient Descent, etc.) [36], [37], [38], [39]. The driving motivation behind this adoption is to proficiently tackle the inherent uncertainties and ambiguities present in real-world scenarios, effectively manage voluminous data, deal with the curse of dimensionality, adapt to dynamic environments, augment the interpretability of models, and improve overall algorithmic performance. With the continued growth in data sizes, the ability to handle big data has become a necessity. The utilization of fuzzy sets helps to manage this data deluge and preempts the exponential explosion of dimensionality. For the sake of giving the readers a comprehensive understanding, the scope of the discussed research herein extends beyond just regression problems, but encompasses a diverse range of machine learning tasks, thereby reflecting the versatility of the incorporation of fuzzy sets in the domain of ensemble and incremental algorithms.

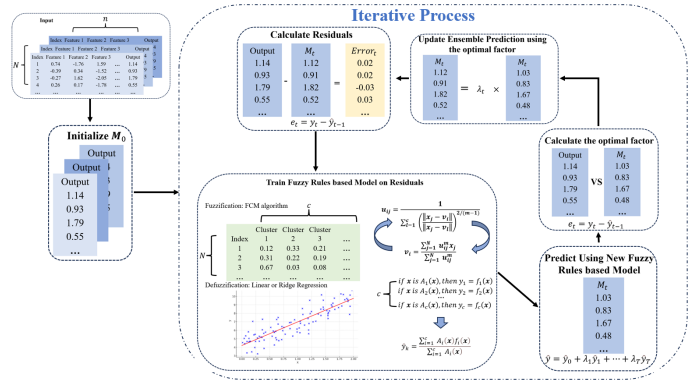

**Fig. 1** Overall structure of the proposed framework

*Enhancements in the Optimization Process*: The authors [19] present HTSK-LN-ReLU, a novel algorithm tackling gradient vanishing in TSK fuzzy systems optimization, enhancing interpretability, robustness, and performance. In [21], the proposed MBGD-RDA algorithm, which merges minibatch gradient descent (MBGD), regularization, and AdaBound, along with three specialized techniques for TSK fuzzy system training, finely tunes Gaussian MFs' parameters and consequent parameters using a regularized loss function for big data regression applications. An interpretable Gaussian Takagi-Sugeno-Kang (TSK) fuzzy classifier, termed WIG-TSK [41], avoids aggregation strategies and permits simultaneous training of all sub-classifiers within a singular structure. The utilization of Gaussian random feature selection to simulate the biological firing pattern of human brain neurons reduce the length of fuzzy rules.

*Approaches to System Integration*: The paper [40] resides in the proposal of a novel method to develop fuzzy rule-based models under the frameworks of both bagging and boosting ensemble mechanisms, and to construct augmented fuzzy rule-based models that amalgamate outputs from diverse models, each developed with varying subsets of data and features. The final output is derived by stacking the outputs of the sub-classifiers, following the proposed minimal-distance-based voting strategy. The framework proposed in [17] employs a TSK fuzzy system fusion at the sensitivity-ensemble-level, measures object sensitivity and implements an ensemble strategy to iteratively under-sample the majority class, thereby preventing overfitting and preserving the contribution of low-sensitive objects. In [42], a stacking-based ensemble Takagi-Sugeno-Kang (TSK) fuzzy classifier, is designed to balance performance and interpretability in high-dimensional feature spaces using statistical information and an entropy-based feature weighting strategy.

*Knowledge Representation and Processing*: In [18], The proposed method integrates a global linear regression sub-classifier with multiple TSK fuzzy sub-classifiers, thereby delivering both feature-importance-based and linguistic-based interpretability. This paper [43] combined fuzzy rule-based models with the theory of deep forest and deep learning-based methods for leaf cultivar classification. This model employs Generative Adversarial Networks (GANs) for contour feature extraction and fuzzy logic for dealing with incomplete and imprecise data. In [44], inspired by human brain knowledge firing patterns, the authors develop the fuzzy rule dropout with dynamic compensation for the Takagi-Sugeno-Kang (TSK) fuzzy classifier, by combing multiple random distributions. Two granular aggregation methods are proposed in [45], based on justifiable granularity, boosting the diversity and interpretability of the fuzzy system. It integrates fuzzy rule-based models in a distributed data environment, forming local models using global data subsets. A comprehensive design for a rule-based model ensemble [46] is introduced, forming a distributed architecture to manage high-dimensional data using an adjustable weight vector, emphasizing the role of correlation strategies in incremental learning.

Despite a wealth of research efforts devoted to augmenting and enhancing fuzzy rule-based models, encompassing the integration of varied model types, the management of high-dimensional data, the incorporation of feature-related processing techniques, and the development of efficient learning methodologies, some investigated algorithms still encounter the following limitations or challenges:

1) Interpretability-Performance Trade-off: Renowned for their interpretability, fuzzy rules-based models have been effectively applied across diverse fields. However, as methodologies become more sophisticated, the performance often improves at the expense of interpretability. For instance, the introduction of deep learning-based models has indeed escalated performance levels. Nevertheless, certain drawbacks associated with deep learning models, such as their inherent black-box nature and the difficulty in understanding their internal workings, have concurrently diminished the interpretability of the model. Consequently, there persists a significant challenge in achieving a harmonious balance between model interpretability and performance.

2) Overfitting: Despite the advent of various strategies aimed at curbing overfitting, this issue remains particularly persistent, especially in scenarios involving high-dimensional data and complex model structures. The need for further advancements in different mechanisms that deter overfitting is imperative to enhance the model's ability to generalize across different data sets.

3) High-Dimensionality: The ability to handle high-dimensional data remains an arduous challenge,

despite numerous methods targeted at enhancing the model's generalizability. The inherent complexity associated with high-dimensional data often leads to overfitting and exacerbates the difficulty in model interpretation.

The proposed approach is underpinned by the fundamental principles of incremental model development. Our framework begins with the simplest and most generalized form one can imagine, serving as our foundational baseline. By leveraging Fuzzy Rule-Based Models as base learners, each iteration retains the interpretability of the model constructed, thereby preserving one of the primary advantages of Fuzzy Rule-Based Models. Simultaneously, through an iterative process, we seek to further diminish the error and enhance the performance of the overall model. Secondly, the framework introduces an optimal factor that ensures the optimal contribution of each model to the overall accuracy and adaptability of the aggregated model. This factor substantially mitigates the risk of premature overfitting, thus enhancing the model's generalization ability. Regarding high-dimensional data, the proposed approach utilizes the inherent structure of the Gradient Boosting technique to handle complexity. Rather than attempting to fit all features simultaneously, the model incrementally learns from the errors of previous learners. This way, even with a high number of features, our framework can achieve an effective and interpretable mapping of the high-dimensional input space to the output. In conclusion, the proposed framework blends the strengths of Gradient Boosting and Fuzzy Rule-Based Models, demonstrating promising potential in overcoming the inherent limitations of traditional fuzzy rule-based models and emerging as an efficacious tool for dealing with regression problems.

## III. AN OVERALL ARCHITECTURE OF THE MODEL

In this study, we present a flexible and adaptable model architecture consisting of a collection of rule-based sub-models (e.g., fuzzy rules-based model). Each of these sub-models is trained and constructed using random subsets of distinct data. Subsequently, we combine these sub-models, adhering to the principle of incremental model development, to gradually improve the overall model's performance. At its core, this design methodology facilitates the effective enhancement of system performance by dividing the system model into smaller, more manageable components. The distinct sub-models can encompass diverse data characteristics, thereby contributing to the enhanced performance of the overall system. In addition, this methodology enables targeted modifications to individual sub-models in response to particular problems or scenarios, thereby enhancing the adaptability of the entire system. The sub-models sequentially execute the following principal processing steps, as depicted in Fig. 1.

We employed Takagi and Sugeno's multiple-input single-output fuzzy system modelling architecture to determine the relationship between the input and output variables in the regression problem. It is necessary to introduce some fundamental notations in this context. This study's experimental data consists of a collection of paired $n$-dimensional input vectors and scalar outputs. Assume we have a finite data set $\boldsymbol{D}$ with $\boldsymbol{N}$ input and output values, denoted as $\boldsymbol{D} = \{(\boldsymbol{x_1}, y_1), (\boldsymbol{x_2}, y_2), \dots, (\boldsymbol{x_N}, y_N)\}$, where $\boldsymbol{x_k} \in R^n$ is the input vector and $y_n \in R$ is the corresponding scalar output.

The development of a "zero model", grounded in the mean output values of the training set, based on output means serves as a starting point, owing to its simplicity and universality, which make it a good baseline model. This choice is predicated on the simplicity and universality of such a model, making it an ideal benchmark for evaluating the performance of more advanced models. By basing the zero model on the training set, we ensure a direct, consistent comparison with the sophisticated models developed within our integrated fusion framework. Because it is independent of input characteristics, this model avoids feature selection and overfitting concerns, resulting in a high degree of generality. Following that, we compute the errors (residuals) between the model and the ground truth and use these residuals to build the fuzzy rule-based model. The fuzzy rule-based model's fundamental goal is to minimize the errors (residuals) caused by the "zero model", hence improving prediction accuracy. To strike a balance between model complexity and generalization capacity, we include a scaling factor, designated as $\lambda_1$, that determines the contribution of the sub model $\boldsymbol{M_1}$. This reduces the difference between the real output and the whole model. Subsequently, we employ an iterative process for the modeling procedure to further diminish the error. For each subsequent sub models (e.g., $\boldsymbol{M_2}, \boldsymbol{M_3}, \dots$), we construct a new rule-based model based on the errors generated by the previous model. Model updating is achieved by accumulating the product of the new model's predictions with the optimized scaling factors into the previous model. This approach allows us to flexibly control the contribution of each sub model in the final integrated model, thus improving prediction accuracy while maintaining a balance between model complexity and generalization performance. To ascertain the comprehensiveness and self-sufficiency of this research, we offer a concise review of the fundamental concepts pertaining to fuzzy rule-based models and gradient boosting techniques. Furthermore, we demonstrate the way this conceptual framework, along with its algorithmic underpinnings, facilitates the integration of multiple sub models into a unified architecture.

### A. The construction of fuzzy model

In this study, we employ Fuzzy C-Means clustering techniques (FCM) [48], [49] and ordinary least squares (OLS) and ridge methods to construct a Takagi-Sugeno (TS) fuzzy model for describing nonlinear systems. The model consists of a set of fuzzy rules, which can be represented as follows:

$$if \; \boldsymbol{x} \; is \; A_i(\boldsymbol{x}), then \; y_i = \alpha_0 + \alpha_1 x_1 + \dots + \alpha_n x_n \qquad (1)$$

Here, $A_i(\boldsymbol{x})$ denotes the membership function of input variable $\boldsymbol{x}$ under the antecedent condition, and $\alpha_0, \alpha_1, \dots, \alpha_n$ represent the consequent parameters based on the fuzzy rule definition. Through this representation, the TS fuzzy rule can capture complex relationships within nonlinear systems, providing a flexible and robust modeling framework.

Considering the capacity of fuzzy C-means clustering (FCM)

to adapt to various shapes and sizes of clusters in the data by adjusting the number of clusters and fuzzification coefficients, thus enabling the model to capture patterns in the data more effectively, as well as its robustness to noise, we employ the FCM clustering algorithm to construct the antecedent part of fuzzy rules. This methodology ensures that the resulting model can handle the inherent complexities and ambiguities of the data while retaining a high level of interpretability and generalizability. The strategy of reducing the complexity and number of base models can enhance the overall understandability of the system. However, it is important to note that when the base models are fuzzy rule bases with clusters in the antecedents and are part of an ensemble, the interpretability may be compromised due to reduced transparency caused by the layered structure and interaction effects. FCM is essentially an iterative optimization of the membership matrix $U = [u_{ij}]$ and clustering centers $v_i$ to minimize the objective function shown below.

$$Q = \sum_{i=1}^{c} \sum_{j=1}^{N} u_{ij}^m \, ||x_j - v_i||^2 \qquad (2)$$

Here $c$ denotes the quantity of clusters, while $v_i$ means the clustering center or prototype for the $i^{th}$ cluster. The fuzzy coefficient, represented by $m$, has an impact on the membership function's form. The notation $||.||^2$ is indicative of the Euclidean distance metric employed in the process. The minimization process entails iteratively updating the partition matrix $U = [u_{ij}]$ and prototypes $v_i$ using the corresponding equations.

$$u_{ij} = \frac{1}{\sum_{l=1}^{c} \left( \frac{||x_j - v_i||}{||x_j - v_l||} \right)^{2/(m-1)}} \qquad (3)$$

$$v_i = \frac{\sum_{j=1}^{N} u_{ij}^m x_j}{\sum_{j=1}^{N} u_{ij}^m} \qquad (4)$$

Given some inputs $x_k$, the following equation could be utilized to determine the corresponding output $\hat{y}_k$.

$$\hat{y}_k = \frac{\sum_{i=1}^{c} A_i(x_k) f_i(x_k)}{\sum_{i=1}^{c} A_i(x_k)} \qquad (5)$$

In the context of fuzzy rule-based models, the consequent parts describe the patterns or trends of the output values when the antecedent conditions are satisfied. These parts are typically represented as linear equations or other forms of functional relationships. A general expression for the consequent part can be formulated as:

$$if \ x \ is \ A_i(x), then \ y_i = \alpha_0 + \alpha_1 x_1 + \cdots + + \alpha_n x_n \qquad (6)$$

Here, $\alpha_0, \alpha_1, ..., \alpha_n$ are model parameters that pertain to the consequent parts in the fuzzy rule-based model definition. The Ordinary Least Squares (OLS) or ridge regression serves as a prevalent technique for determining the consequent parameters. In the realm of fuzzy rule-based models, zero-order, first-order, and second-order fuzzy rule-based models are widely employed in diverse applications, primarily differentiated by the way input variables are incorporated into the consequent parts. Zero-order models exclude input variables, first-order models encompass first-degree terms of input variables, and second-order models integrate both first-degree and second-degree terms of input variables. These models exhibit a spectrum of complexity and fitting capabilities; higher-order models possess the capacity to capture more sophisticated data patterns, albeit with an increased risk of overfitting.

*B. Gradient boosting for incremental learning.*

The incremental principle facilitates a gradual refinement of the problem-solving approach, thereby mitigating the challenges associated with voluminous and complex models. Initially, at step $t_0$, a "zero" model, represented by the mean of outputs, serves as the starting point for this model. Throughout each iteration ($t = 1, 2, ..., T$), the process begins by computing the error between the predicted outputs and the actual outputs based on the current model.

$$e_t = y_t - \hat{y}_{t-1} \qquad (7)$$

A new fuzzy rule-based model is then fitted by minimizing this error. Fundamentally, each iterative step in the modeling process aims to further reduce the error. Given an input $x$, the final output of the ensemble model can be expressed as follows:

$$\hat{y} = \hat{y}_0 + \hat{y}_1 + \cdots + \hat{y}_T \qquad (8)$$

Here, $\hat{y}_0, \hat{y}_1, ..., \hat{y}_T$ are the output of the model $M_0, M_1, ... M_T$. It is worth noting that the number of rules for each newly constructed model may vary depending on the nature of the error and the complexity of the dataset. For instance, when employing linear functions, three groups or rules might suffice. However, in more complex scenarios, it may be necessary to adopt a greater number of rules, such as four, five, or even more. While increasing $c$ can enhance the performance of rule-based models, excessive optimization may lead to overfitting or "memorization" of the data. Consequently, a balance between model complexity and generalization ability for new data must be struck.

*C. Contribution of each fuzzy rule-based model*

In our study on the Gradient Boosting ensemble algorithm, it is critical to ensure a degree of diversity among the base models. This diversity serves to enhance the breadth of information sourced during model integration, thereby boosting the model's generalization capabilities. In the proposed methodology, the diversity of models is achieved not only through learning from residuals but also by constructing distinct fuzzy rule-based models at each iteration. These models are tailored to discern the relationship between current data features and residuals. However, it is simplistic and potentially inaccurate to assume an equal contribution from each model towards the collective performance of the overall ensemble. As the number of models in the ensemble grows, the novelty of information unearthed by each newly added model is likely to vary. With this consideration in mind, we introduce a factor to regulate the contribution of the current model to the overall ensemble. Essentially, this factor dictates the learning pace of the whole ensemble model. The optimization of this factor is achieved through the execution of a grid search algorithm on a validation set. The output of the overall ensemble model can be calculated with the following formula.

$$\hat{y} = \hat{y}_0 + \lambda_1 \hat{y}_1 + \cdots + \lambda_T \hat{y}_T \qquad (9)$$

The design of our algorithm optimally balances diversity and the individual contribution of each model, leading to enhanced model performance. Our findings highlight the efficacy of integrating Gradient Boosting with fuzzy rule-based models and the potential benefits of adjusting the contribution of different models in ensemble learning. In our study, we opt for the Root Mean Square Error (RMSE) as the performance metric for our proposed algorithm. The RMSE is a widely used and robust measure of the differences between values predicted by a model and the actual observed values. It can be computed via the subsequent equation provided below.

$$RMSE = \sqrt{\sum_{i=1}^{N} \frac{(\hat{y}_i - y_i)^2}{N}} \qquad (10)$$

Reliant on the performance feedback garnered from a validation set, which constitutes half the size of our training set, we can make appropriate adjustments to the complexity of our model. For example, should the model exhibit poor performance on the validation set, this could indicate a need to simplify the model to mitigate the risk of overfitting to the training data. It provides an unbiased environment to evaluate the model's generalization capabilities, ensuring that our model's complexity is optimally tuned not just to fit the training data, but also to perform well on unseen data. Consequently, this would necessitate simplification of the model, reducing its complexity to better generalize from our training data to unseen data, hence minimizing the risk of overfitting.

---

**Algorithm 1** Fuzzy Systems in Gradient Boosting Framework

**Input:** $D = \{(x_1, y_1), (x_2, y_2), \ldots, (x_N, y_N)\}$ where $x_k \in R^m$ is the input vector and $y_n \in R$ is the corresponding scalar output.
Tolerance $\varepsilon$: a small value used to decide when to stop adding sub models.
The number of iterations $T$.
**Output:** The final model $M_t$
**Function** computeResidual $(y, M_{t-1}(x))$:
  $i \leftarrow 1$
  **repeat**
    $e_{ti} \leftarrow y_i - M_{t-1}(x_i)$
    $i = i+1$
  **until** $i = N$
  return $e_t$
**Function** fuzzyCMeansCluster($x, c$):
  Initialize the membership matrix $U$ randomly
  **repeat**
    Calculate cluster centers $v$ based on $U$
    Update $U$ based on $v$ and $x$
  until convergence criteria met
  return $U, v$
**Function** fuzzyRuleBasedModel($x_i, c$, LinearModels):
  Initialize output $y_i \leftarrow 0$
  **for** each cluster $j$ from 1 to $c$
    Calculate membership $u_{ij}$ for input $x_i$ using cluster center $v_j$
    Calculate local model output $y_{ij}$ using LinearModels[j] and $x_i$
    $y_i = y_i + u_{ij} y_{ij}$
  **end** for
  return $y_i$
Main Algorithm:
  Initialize the "zero model" $M_0$ as the mean of the output values from the training dataset:
  **do**
    $e_t \leftarrow$ computeResidual $(y, M_{t-1}(x))$:
    Train fuzzy rule-based model $M_t(x)$ using $\{(x_i, e_{ti}), i = 1$ to $N$
    Compute the scaling factor $\lambda_t \leftarrow argmin \sum |y_i - (\hat{y}_{i(t-1)} + \lambda_{t-1} M_{t-1}(x_i))|$ for all $(x_i, e_{ti})$ .
    Update the model: $\hat{y}_{it} = \hat{y}_{i(t-1)} + \lambda_t M_t(x_i)$
  until stop criteria met
  The final model is given by: $\hat{y}_{iT} = M_T(x_i)$

---

Algorithm 1 delineates the comprehensive pseudocode of our approach, encapsulating the pivotal stages of the framework, including initial model estimation, residual computation, and model iteration among others. The framework, from a theoretical standpoint, adheres to the principle of incrementality in model development [50]. The initial phase of our approach, termed as the 'zero model', generated from the training set, serves as a simplistic estimator and lays down a baseline for regression tasks. The zero model, devoid of any presumptions about the underlying distribution or relationship between input and output variables, showcases its simplicity and universality by providing a solid foundation for enhancing subsequent model performance. The average computational complexity of this stage is O($N$), with $N$ being contingent on the size of the dataset. The cornerstone strategy of this framework is the construction of models predicated on the residuals of the preceding model. An iterative process of fitting new rule-based fuzzy models to these residuals is deployed, and after each iteration, the overarching model is updated. This incremental learning method proves efficacious in reducing bias, particularly in dealing with complex datasets exhibiting high non-linearity, at the cost of an increase in computational complexity with each iteration. For each fuzzy rule-based model, the Fuzzy C-Means (FCM) method is applied to partition the input space into distinct regions, each of which can be represented by a fuzzy rule. The time complexity for FCM is typically O($qcN$), where $q$ denotes the number of iterations, $c$ represents the number of clusters, and $N$ stands for the number of data points. If the number of dimensions is high, it could substantially escalate the time complexity due to the calculation of distances between data points and cluster centroids in multi-dimensional space in each iteration. Upon generation of fuzzy rules, linear regression is utilized to optimize the parameters of these rules, which is generally associated with a time complexity of O($Nmp^3$), where $m$ refers to the number of rules and $p$ denotes the number of regressors in each rule. In summation, in contrast to conventional gradient boosting models, this approach may necessitate increased computational resources due to the incorporation of fuzzy rule-based models. However, the interpretability of fuzzy systems and their robustness to noise and outliers lend them the potential to offer superior performance on complex real-world datasets. This approach also endows the model with interpretability, as each rule can be scrutinized and comprehended by human experts. Since this is an iterative and incremental process, the number of iterations and the quantity of rules in each fuzzy rule-based model will influence the model's learning time and accuracy. This trade-off between complexity and performance is a prevalent consideration in machine learning model design.

## VI. EXPERIMENTS AND RESULTS ANALYSIS

In this section, we present a comprehensive experimental evaluation of our proposed method, which employs the concept of gradient boosting to integrate models based on fuzzy rules, aiming to measure the performance of the implemented methodology.

### A. Synthetic and real-world datasets

Initially, we elaborate the entire process of the algorithm by using a synthetic dataset, demonstrating the underlying mechanism in a clear and straightforward manner. We constructed the synthetic dataset by generating 4,000 linearly spaced values, $x$, from 1 to 4000, and calculating corresponding $y$ values using $y = sin(x) + \sqrt{x/2} + exp(x/15)$. For each fuzzy rule-based model, there exist two critical parameters to consider: the number of rules and the fuzzification coefficients. In the context of the experiments performed, the number of rules varied within the range of 2 to 10. On the other hand, the fuzzification coefficients underwent alterations within the range of 1.1 to 2.9. During the ensemble process, the maximum number of iterations was set to 100.

As discussed in the preceding sections, our point of departure is a model referred to as $M_0$, a model of constancy which, in our conception, is synonymous with an average value. This average value acts as a fundamental baseline from which our boosting algorithm begins. Subsequently, the discrepancy between the real values and the output of $M_0$, a quantifiable error, becomes the target for the creation of our next model, $M_1$. It's worth noting that our perspective of the subsequent model $M_1$ is an error compensator for the model $M_0$. Put differently, our intent is to have the rule-based model $M_1$ to compensate for the difference between the actual values and the outputs from the $M_0$. Therefore, $M_1$'s role is essentially to estimate and compensate for these errors. Following the establishment of $M_1$, we continue this iterative process by introducing the subsequent model, $M_2$. Like its predecessor, $M_2$ is designed to serve as an error estimator and compensator, but this time its task is to address the discrepancy between the actual values and the combined outputs from $M_0$ and $M_1$. The fundamental philosophy of this process remains unchanged: each new model, from $M_1$ through to $M_T$, is essentially a compensator and corrector, each attempting to redress the discrepancy, or error, between the real values and the output of the ensemble model that preceded it. The iterative development of these models is a manifestation of the underlying gradient boosting concept in our method, with each model serving as a step along the gradient descent towards the optimization goal. As we advance through this iterative process, with each model that is introduced, the output of the overall ensemble model improves, becoming more aligned with the actual values. $M_T$, where T represents the final stage of the iterative process, is the culmination of this series of improvements. $M_T$ estimates and compensates for the differences between the actual values and the output of the ensemble model $M_{T-1}$.

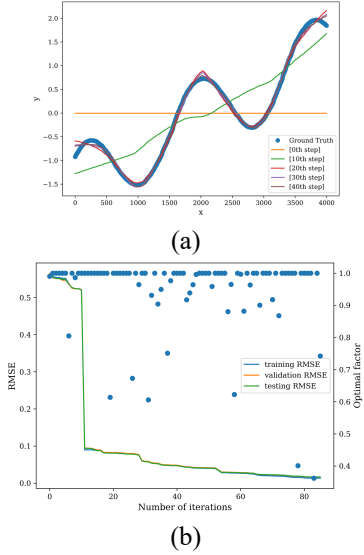

(a)

(b)

**Fig. 2.** (a) Comparative Visualization of the Model Output at Various Stages Against the Ground Truth; (b) Illustration of the Root Mean Square Error (RMSE) Evolution Corresponding to Iteration Changes and the Optimal Factor.

Referencing Fig. 2 (a) which illustrates the output trajectory of the ensemble model at various stages of its incremental development, a discernible pattern emerges regarding the discrepancy between the actual values and the predictions produced by our model. As the process progresses and an increasing number of fuzzy rule-based models are integrated into the ensemble model, the divergence between these actual and predicted values manifests a notable trend of reduction. As a result, shown in Fig. 2 (b), this iterative process yields a systematic diminution of the gap between the actual and predicted values, indicating an ongoing improvement in the predictive accuracy of the ensemble model. The continual contraction of this discrepancy is a testament to the efficacy of our proposed gradient boosting approach in enhancing the performance of fuzzy rule-based models. Not only does it validate our methodology but also elucidates the potential benefits of integrating gradient boosting with fuzzy rule-based systems in creating an adaptive, flexible, and efficient learning model.

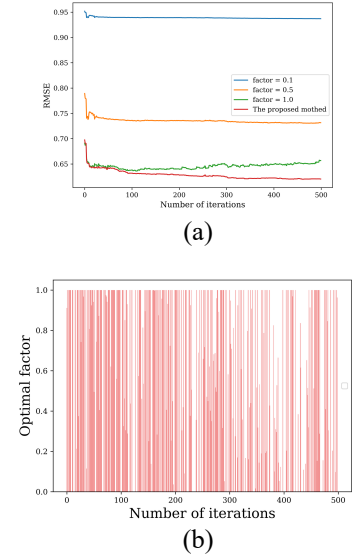

(a)

(b)

**Fig. 3.** (a) Comparative Analysis of Fixed Versus Dynamic Contribution Approaches; (b) Depiction of Optimal Factors Across Varied Iterative Stages.

Moreover, during this incremental learning process, we introduce a control factor, lambda ($\lambda$), into the modeling. This parameter essentially dictates the contribution of each newly incorporated model to the collective model. We see lambda as a modulating factor, allowing us to control and adjust the influence of each individual model on the overall system, thus providing a degree of flexibility and adaptability in the learning process. This not only refines the model but also ensures that each incremental step contributes optimally to the improved accuracy of the ensemble. As shown in Fig. 2 (b), it is crucial to re-emphasize the role of the control parameter lambda (λ) at this point. Lambda plays a pivotal role in controlling the contribution of each new model, starting from $M_1$ and continuing through to $M_T$, adjusting the influence of each individual model on the overall system. Through this control mechanism, the ensemble system remains adaptable and flexible, capable of adjusting according to the complexity of the data and the task at hand.

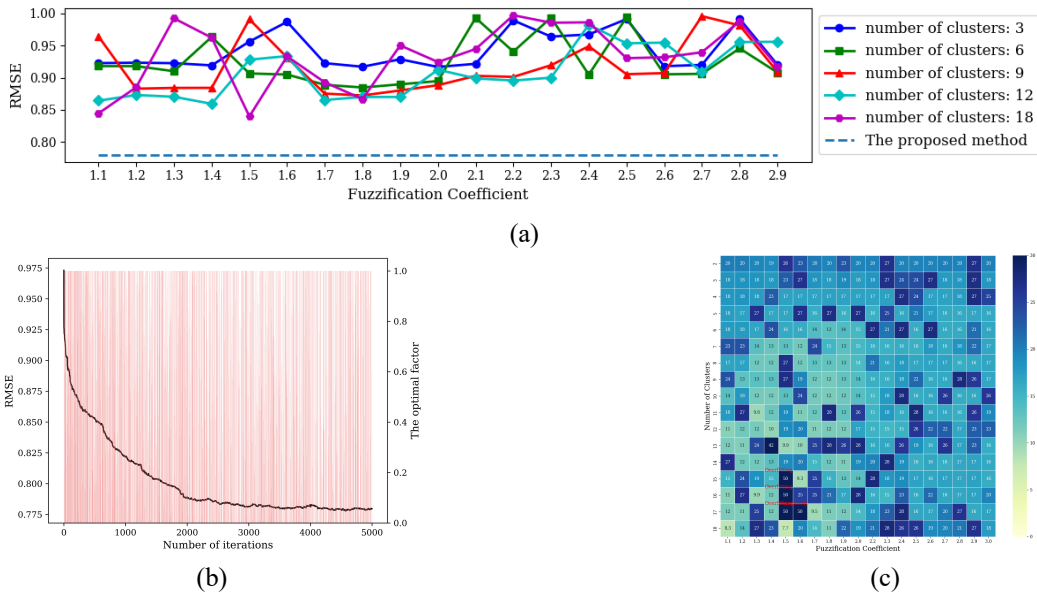

(a)

(b)                                                             (c)

**Fig. 4** Comparison of the proposed method vs. fuzzy rule-based models across multiple parameter settings.

To further elucidate the pivotal role of the optimal factor in ensuring the optimal contribution of each step to the overall model's accuracy and adaptability, we carried out an analysis leveraging conventional, static contribution values, specifically 0.1, 0.5, and 1.0, on the widely used real-world Alabone dataset sourced from the UCI Machine Learning Repository [52][53]. These constant contribution values imply that each model has a uniform impact during the ensemble process. Keeping all other parameters constant, we present a comparative assessment of Root Mean Square Error (RMSE) at each gradient boosting step in Fig. 3. The RMSE, serving as an indicator of model performance, provides insights into the effect of varying contribution factors on model accuracy and adaptability.

From a preliminary glance at the figure, it is evident that a smaller factor value corresponds to a reduced contribution from the model, resulting in a slower decline in the test RMSE. On the contrary, a larger factor value means a greater contribution from the model, leading to a more rapid reduction in the test RMSE. However, this accelerated decrease in RMSE comes at the cost of an increased risk of overfitting. This risk becomes particularly prominent when the factor equals 1, where overfitting indeed manifests itself. However, in our proposed methodology, we employ a validation set during the gradient boosting process to tune the factor dynamically. This strategy effectively mitigates the risk of overfitting, as it continuously adjusts the factor based on the model's performance on the validation set. This adaptive approach enables the system to balance the contributions of individual models more effectively, facilitating an optimal trade-off between model accuracy and the risk of overfitting. Consequently, it underscores the importance of tuning the factor dynamically in the context of gradient boosting, particularly when integrating fuzzy rule-based models in an ensemble.

In a rigorous endeavor to elucidate the merits of the principle of incremental model development, we performed a comprehensive suite of experiments that systematically varied key parameters. Utilizing the Park Motor UPDRS dataset [52 - 53]—a benchmark in real-world applications—as our empirical backdrop, we juxtaposed the efficacy of our newly proposed methodology with that of fuzzy rule-based models. Two pivotal parameters, namely the number of clusters and fuzzification coefficients, exert a significant influence on the performance landscape of fuzzy rule-based models. For the former, we undertook an empirical investigation that spanned a gamut of cluster counts, ranging from 2 to 18. In each experimental iteration, we incremented the cluster count by a single unit, thereby furnishing insights into the model's adaptability across a spectrum of data complexities—from simplistic to intricate scenarios. With respect to fuzzification coefficients, an element integral to the determination of membership gradations, we executed a parametric sweep from 1.1 to 2.9, incremented at intervals of 0.1. This selection criterion was strategically designed to not only cover a range of commonly employed fuzzification paradigms but also to probe the resilience and adaptability of the model under extremities of parameter settings. To ensure methodological coherence, the parameter settings for our proposed method were kept identical to those used in the experiments described above.

Figure 4(a) elucidates the Root Mean Square Error (RMSE) of fuzzy rule-based models across an array of fuzzification coefficients. Within this illustration, various lines symbolize distinct cluster counts, challenging the premise that an increment in cluster quantity invariably elevates the model's predictive accuracy. The hypothesis of performance enhancement primarily credits the Fuzzy C-Means (FCM) algorithm's aptitude for segmenting the data space into a wider array of fuzzy subsets. Such escalation in the number of clusters facilitates the model in encapsulating more intricate and subtle data patterns, thereby enriching the rule generation process. Nonetheless, the anticipated trend is not discernible in Figure 4(a). For instance, the outcome with nine clusters (indicated by the red line) does not demonstrate a significant improvement over the model with three clusters (represented by the blue line). This figure unveils a nonlinear and inconsistent correlation between the cluster numbers and RMSE across varied fuzzification coefficients, indicating that merely augmenting the number of clusters does not assuredly amplify model performance. This complexity underscores a nuanced interdependence between the number of clusters and fuzzification coefficients, critically impacting the model's proficiency in identifying underlying data patterns, especially in complex datasets. Figure 4(b) illustrates that our proposed methodology, employing a gradient boosting strategy to incrementally integrate additional fuzzy rule-based models, exhibits a significant declining trend in the Root Mean Square Error (RMSE) on the test set as more models are combined. In comparison to a singular fuzzy rule-based model, a lesser quantity of clusters may result in insufficient capability to capture the intrinsic information within the data, thereby highlighting performance limitations. However, with a substantial number of clusters, traditional rule-based fuzzy models reach a certain performance plateau, until the inherent characteristics of the data, such as noise and outliers, lead to noticeable overfitting. The proposed method circumvents these pitfalls by adhering to an incremental development principle.

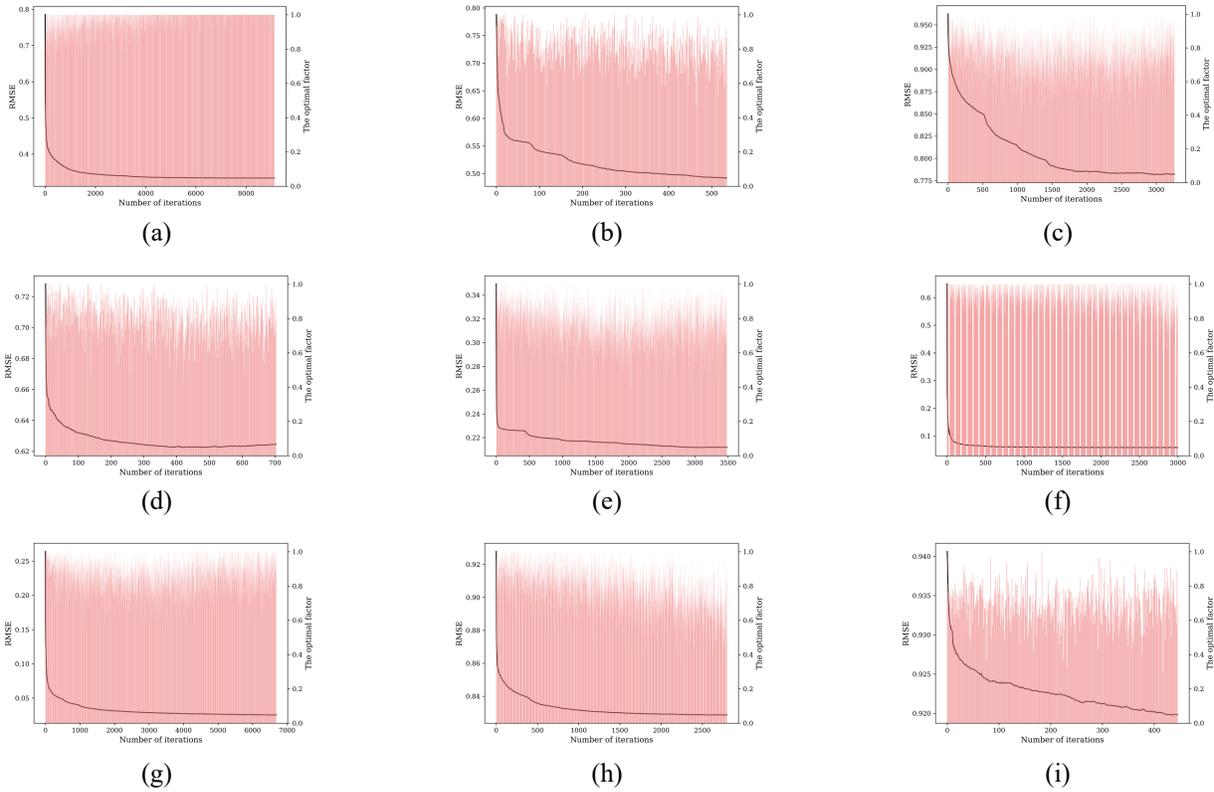

**Fig. 5** Experimental Results and Optimal Factors for (a) Scikit-digits; (b) Space GA; (c) Park Motor UPDRS; (d) Abalone; (e) Power Plant; (f) Naval; (g) Steel Industry; (h) Year Prediction MSD; (i) Microsoft.

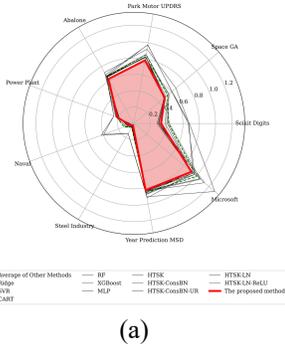

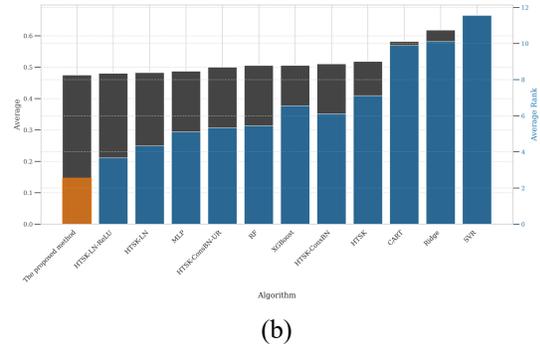

**Fig. 6** Comparison of Different Algorithms on multiple datasets.

By synergistically combining different fuzzy rule-based models within a gradient boosting framework, we capture more complex data patterns, mitigate the risk of overfitting, and diminish the dependence of a single model on initial conditions. In Figure 4(c), we quantify the performance improvement of our new algorithm over traditional fuzzy rule-based models with varying parameters. The results show a significant performance enhancement of 18.55%±6.25% compared to traditional models under different parameters. This demonstrates that the proposed method not only enhances model accuracy but also ensures a more robust and reliable prediction performance, illustrating the effectiveness of integrating multiple models to address the complexities inherent within dataset.

TABLE I
SUMMARY OF THE REAL-WORLD DATASETS.

| | Number of instances | Features | Dataset Size (Training) | Dataset Size (Training) |
|---|---|---|---|---|
| Scikit-digits | 1,797 | 64 | 1,258 | 539 |
| Space GA | 3,107 | 6 | 2,174 | 933 |
| Park Motor UPDRS | 5,875 | 16 | 4,112 | 1,763 |
| Abalone | 4,177 | 8 | 2,923 | 1,254 |
| Power Plant | 9,568 | 4 | 6,697 | 2,871 |
| Naval | 11,934 | 16 | 8,353 | 3,581 |
| Steel Industry | 35,040 | 9 | 24,528 | 10,512 |
| Year Prediction MSD | 51,534 | 90 | 36,073 | 15,461 |
| Microsoft | 49,862 | 136 | 34,903 | 14,959 |

In a bid to further analyze the performance and efficacy of the proposed algorithm, we have implemented it on several real-world datasets derived from the UCI Machine Learning Repository [52] [53]. These datasets exhibit diversity in terms of sample size and dimensions, allowing for a comprehensive evaluation of our algorithm across varying data landscapes. All other parameters remain consistent with those used for the synthetic dataset. However, for larger datasets like the Year Prediction MSD dataset, we expand the range of the number of rules in fuzzy rule-based models from [2,10] to [2,20] and increase the maximum number of iterations to 10,000. This adjustment is necessitated by the fact that as the dimensions and the number of samples in a dataset increase, the feature space correspondingly becomes more diverse and potentially complex. To ensure a thorough and accurate depiction of this expanded and more intricate feature space, it becomes essential to define a larger quantity of rules. In other words, in the face of increasingly intricate decision boundaries or relationships within the data, our model requires a significant number of rules and robust fuzzy rule-based models to effectively capture these sophisticated relationships. Table I provides the description of each dataset, showing the division of training and testing datasets.

The proposed algorithms are evaluated with a couple of existing algorithms, encompassing six classical machine learning algorithms as well as five recent gradient and/or Takagi-Sugeno-Kang (TSK) related models derived from [19], [20], [51]. Table II meticulously outlines the configuration of parameters for each algorithm under consideration, ensuring a thorough understanding of experimental results. Specifically, for the recent gradient and/or TSK-related models, the Gaussian Membership Functions (MFs) utilized in these models play a pivotal role in their performance. Two critical parameters for these MFs—the standard deviation values and the centers of the antecedent Gaussian MFs—are initialized at $\{0.1, 0.3, 0.5, 0.7, 1.0\}$ and via k-means clustering, respectively. Additionally, the determination of the number of Gaussian MFs (K) within each input domain is critical for tailoring the model's complexity to the specific characteristics of the data, with values $\{500, 1500, 3000\}$ being explored to balance model accuracy and computational efficiency. Key parameters such as minibatch (BN) sizes (set at $\{64, 512\}$), the DropRule rate (0.5), the initial learning rate (0.01), and the regularization coefficient (0.05) have been carefully chosen to enhance the performance of the algorithms. These parameters play integral roles in managing the learning process, preventing overfitting, and ensuring a robust generalization capability across diverse datasets. Additionally, incorporating techniques such as first layer normalization (LN), uniform regularization (UR), layer normalization (LN), and rectified linear unit (ReLU) can improve the performance of the method.

Table III and Figures 5-6 present a detailed comparison of performance metrics across different algorithms. Figure 5 illustrates the change of the Root Mean Square Error (RMSE) at each iteration stage, along with the corresponding optimal factor that governs the contribution of the model generated at step $i$th. From the experimental results, we observe several common behavioral patterns: (i) With the ensemble's evolution, incorporating more models, subsequent additions tend to contribute less to reducing the RMSE. In other words, model integration later in the process typically offers smaller performance enhancements. This phenomenon aligns with the concept of information saturation: as the number of models increases, each new model may discover less new information compared to its predecessor. Consequently, the incremental performance improvement provided by each subsequent model may gradually diminish; (ii) During the ensemble process, individual models exhibit varying degrees of influence on the overall ensemble performance. This variation can be attributed to the unique characteristics of the data or may reflect the models' differing capabilities in capturing underlying data patterns. Figure 6(a) visualizes the RMSE values for all methods across all datasets, where the red region indicates the areas where the proposed method outperforms other approaches. Figure 6(b) lists the average RMSE values for all methods across different datasets, corresponding to the black bars and the scale on the left y-axis. The orange (the proposed method) and blue (the other methods) bars represent the average performance ranking of different algorithms across datasets. It is observed that both the average ranking and the average RMSE values are lower for the proposed algorithm than for other methods, indicating a certain superiority of our approach. Further, we employed a *t*-test to analyze these results, which revealed statistically significant performance differences between the proposed method and common methods (e.g., SVR with p = 0.00115, XGBoost with p = 0.00834). In comparisons with related proposed models (e.g., HTSK-ConsBN-UR with p = 0.09596), while the performance of our method appears superior in some cases, these differences did not reach statistical significance (p >= 0.05). The absence of statistical significance in these comparisons can be attributed to several factors, such as the use of FCM, which may not cluster complex datasets effectively. Poor clustering results might limit the discernibility of performance improvements. Here, we limited ourselves to using a similar type of base learner, fuzzy rule-based model, not fully leveraging the potential of gradient boosting to enhance overall model results through diversity. Considering this, exploring different strategies to construct fuzzy rule-based models and experimenting with a broader range of base learners represents a direction for our future research.

TABLE II
ENUMERATION OF COMPARATIVE ALGORITHMS
WITH THEIR RESPECTIVE PARAMETER
CONFIGURATIONS.

| Algorithm | Parameter Settings |
|-----------|-------------------|
| Ridge | Weight of L2 regularization adjusted across $\{0.01, 0.1, 0.1, 10, 100\}$. |
| Support Vector Regression (SVR) | Regularization parameter C adjusted across $\{0.01, 0.1, 0.1, 10, 100\}$. |
| Classification and Regression Trees (CART) | Maximum tree depth adjusted across $\{3, 5, 7, 9\}$. |
| Random Forest (RF) | Maximum tree depth adjusted across $\{3, 5, 7, 9\}$. |
| XGBoost | Maximum tree depth adjusted across $\{3, 5, 7, 9\}$. |
| Multilayer Perceptron (MLP) | A fully connected neural network and Adam was used as the optimizer. Same parameter settings as [19] |
| HTSK | Same parameter settings as [20] |
| HTSK-ConsBN | Same parameter settings as [51] |
| HTSK-ConsBN-UR | Same parameter settings as [51] |
| HTSK-LN | Same parameter settings as [19] |
| HTSK-LN-ReLU | Same parameter settings as [19] |

*B. Experimental observations and discussion*

In light of our experimental findings, we propose a multi-faceted discussion and analysis to thoroughly comprehend the implications of the results and lay the groundwork for further enhancements to our methodology.

In this research, we present an approach grounded in the principle of incremental model development, drawing on the power of ensemble learning to mitigate errors related to bias and variance. The development commences with the simplest and most generalized form, the average estimator, reflecting the foundational philosophy of minimizing over-reliance on individual models and side-stepping their inherent limitations. Additionally, the incorporation of gradient boosting techniques serves as a pivotal component in the proposed methodology. By systematically concentrating on the residuals produced by preceding models, the process effectively refines subsequent models to adeptly predict intricate, nonlinear relationships. Simultaneously, the complexity of the overall model is dynamically adjusted based on its performance within a validation set, thereby reducing the risk of overfitting. This dynamic adjustment of complexity plays an essential role in ensuring that the model is robust yet flexible. Moreover, the integration of fuzzy rule-based models within the ensemble significantly enhances the system's interpretability. At each iterative step, the generation of a relatively limited number of fuzzy rules facilitates systematic understanding and analysis. To an extent, each rule within the fuzzy rule-based model is accessible for examination by human experts, thereby enabling a more transparent interpretation of the information contained within the dataset. This approach not only contributes to model's performance but also aligns with an emerging paradigm in machine learning that emphasizes transparency and explicability. By making the underlying rules and relationships accessible and comprehensible, it allows for the possibility of scrutiny by human experts, thereby facilitating a more profound understanding of the information encapsulated within the dataset.

While our method has outperformed other comparative models in terms of performance index and rankings, it is crucial to acknowledge certain limitations that can be addressed in future work: (i) Computational Complexity: Given the iterative nature of our approach, and the necessity of generating and integrating multiple fuzzy rule-based models, the computational overhead can be substantial, especially for larger datasets. (ii) Model Contribution Variance: We observed that, during the ensemble process, individual models exhibited varying degrees of influence on the overall ensemble model. This variance, while potentially reflective of the data's unique characteristics, could lead to inconsistency in the contribution of different models, which could affect the robustness of the

TABLE III

PERFORMANCE METRICS: RMSES FOR DIFFERENT ALGORITHMS OVER DATASETS.

| | Scikit Digits | | Space GA | | Park Motor UPDRS | | Abalone | | Power Plant | |
|---|---|---|---|---|---|---|---|---|---|---|
| | Mean | STD | Mean | STD | Mean | STD | Mean | STD | Mean | STD |
| Ridge | 0.6658 | 0.0000 | 0.6060 | 0.0000 | 0.9763 | 0.0000 | 0.6695 | 0.0000 | 0.2619 | 0.0000 |
| SVR | 0.6787 | 0.0000 | 0.6077 | 0.0000 | 0.9731 | 0.0000 | 0.6892 | 0.0000 | 0.2620 | 0.0000 |
| CART | 0.6751 | 0.0000 | 0.6715 | 0.0000 | 0.9216 | 0.0000 | 0.7176 | 0.0000 | 0.2380 | 0.0000 |
| RF | 0.4084 | 0.0068 | 0.5622 | 0.0028 | 0.8162 | 0.0046 | 0.6683 | 0.0033 | 0.1928 | 0.0009 |
| XGBoost | 0.3805 | 0.0123 | 0.5522 | 0.0090 | 0.8561 | 0.0061 | 0.6718 | 0.0085 | 0.1893 | 0.0062 |
| MLP | 0.2815 | 0.0148 | 0.4893 | 0.0120 | 0.7856 | 0.0183 | 0.6359 | 0.0083 | 0.2315 | 0.0015 |
| HTSK | 0.4332 | 0.1743 | 0.4985 | 0.0110 | 0.8289 | 0.0081 | 0.6646 | 0.0085 | 0.2230 | 0.0019 |
| HTSK-ConsBN | 0.3091 | 0.0483 | 0.4958 | 0.0103 | 0.8245 | 0.0075 | 0.6589 | 0.0080 | 0.2240 | 0.0014 |
| HTSK-ConsBN-UR | 0.2996 | 0.0230 | 0.5108 | 0.0143 | 0.8252 | 0.0096 | 0.6546 | 0.0081 | 0.2216 | 0.0010 |
| HTSK-LN | 0.3122 | 0.0326 | 0.4795 | 0.0092 | 0.8276 | 0.0137 | 0.6520 | 0.0074 | 0.2240 | 0.0013 |
| HTSK-LN-ReLU | 0.2979 | 0.0077 | 0.4781 | 0.0114 | 0.8257 | 0.0183 | 0.6560 | 0.0062 | 0.2211 | 0.0015 |
| The proposed method | 0.3338 | 0.0033 | 0.4914 | 0.0023 | 0.7825 | 0.0014 | 0.6251 | 0.0009 | 0.2118 | 0.0002 |

| | Naval | | Steel Industry | | Year Prediction MSD | | Microsoft | | Average | Average Rank |
|---|---|---|---|---|---|---|---|---|---|---|
| | Mean | STD | Mean | STD | Mean | STD | Mean | STD | | |
| Ridge | 0.3938 | 0.0000 | 0.1400 | 0.0000 | 0.8808 | 0.0000 | 0.9711 | 0.0000 | 0.6184 | 10.1111 |
| SVR | 0.4263 | 0.0000 | 0.1419 | 0.0000 | 0.9105 | 0.0000 | 1.2943 | 0.0000 | 0.6649 | 11.5556 |
| CART | 0.1039 | 0.0000 | 0.0467 | 0.0000 | 0.9247 | 0.0000 | 0.9408 | 0.0000 | 0.5822 | 9.8889 |
| RF | 0.0734 | 0.0031 | 0.0335 | 0.0007 | 0.8788 | 0.0008 | 0.9177 | 0.0006 | 0.5057 | 5.4444 |
| XGBoost | 0.0881 | 0.0032 | 0.0384 | 0.0048 | 0.8577 | 0.0011 | 0.9180 | 0.0018 | 0.5058 | 6.5556 |
| MLP | 0.1070 | 0.0142 | 0.0537 | 0.0016 | 0.8310 | 0.0039 | 0.9711 | 0.0250 | 0.4874 | 5.1111 |
| HTSK | 0.0760 | 0.0435 | 0.0355 | 0.0023 | 0.8454 | 0.0055 | 1.0621 | 0.0964 | 0.5186 | 7.1111 |
| HTSK-ConsBN | 0.0787 | 0.0221 | 0.0429 | 0.0024 | 0.8362 | 0.0036 | 1.1274 | 0.2867 | 0.5108 | 6.1111 |
| HTSK-ConsBN-UR | 0.0597 | 0.0040 | 0.0438 | 0.0022 | 0.8360 | 0.0032 | 1.0525 | 0.1212 | 0.5004 | 5.3333 |
| HTSK-LN | 0.0375 | 0.0041 | 0.0298 | 0.0014 | 0.8406 | 0.0042 | 0.9435 | 0.0219 | 0.4830 | 4.3333 |
| HTSK-LN-ReLU | 0.0313 | 0.0034 | 0.0287 | 0.0013 | 0.8407 | 0.0029 | 0.9471 | 0.0160 | 0.4807 | 3.6667 |
| The proposed method | 0.0582 | 0.0000 | 0.0253 | 0.0003 | 0.8286 | 0.0004 | 0.9198 | 0.0003 | **0.4752** | **2.5556** |

overall ensemble. (iii) Information Saturation: As more models are incorporated into the ensemble, each additional model tends to contribute less to the performance improvement, which could lead to diminishing returns over time. (iv) While our framework shows promise, its performance is closely tied to the choice of base learners. While using Fuzzy Rule-Based Models has its merits, there may be scenarios where alternative base learners could offer better performance or efficiency. This dependency underscores the need to further investigate and benchmark different base learners within our framework.

## V. CONCLUSION

In this paper, we provide an innovative methodological framework for ensemble learning, based on Gradient Boosting techniques and Fuzzy Rule-Based Models as the base learners. The framework's unique configuration and control factor utilization have proven to be effective strategies against overfitting, a persistent challenge in machine learning. The control factor plays an import role, ensuring that no individual model dominates during the training phase, limiting model complexity, and further provides a dynamic tuning mechanism, protecting against overfitting due to data imbalances or noise. The experimental results underscore the efficacy of the proposed framework, demonstrating considerable performance enhancement. Given the inherent flexibility of the proposed ensemble learning framework, future research can explore its expansion and generalization capabilities. A promising direction lies in investigating the feasibility and effectiveness of replacing the base learner with alternative models or algorithms. The performance of different base learners within this framework can be benchmarked and compared, providing a rich landscape for exploration and optimization in ensemble learning methods. Additionally, due to the computational requirements associated with the Fuzzy C-Means (FCM) algorithm, which is employed in establishing the antecedent part of the fuzzy rule-based model, future work could focus on streamlining this process. The computational efficiency of the iterative construction of fuzzy rule-based models within the framework presents a crucial area for improvement. Thus, the development and incorporation of methods to expedite the creation of these models at each iteration is a valuable direction for future research. In addition to the aforementioned, we aim to further enhance the interpretability of the model. Our objective is to augment the system's overall interpretability without compromising its performance. We contemplate providing users with tools and methodologies to facilitate a more profound understanding of the model's operational mechanics. This might entail the development of pertinent visualization tools for better elucidation of the model's inner workings.

**Jinbo Li** received his Ph.D. in Software Engineering and Intelligent System from the Department of Electrical and Computer Engineering at the University of Alberta, Edmonton, Canada, in 2018. He specializes in the development and application of advanced computational methodologies, with a particular focus on Fuzzy Artificial Intelligence Systems, Time Series Anomaly Detection, and Data Mining. His scholarly pursuits also extend to the realm of Interpretable Fuzzy Systems and Deep Learning architectures. He has contributed to multiple peer-reviewed journals and conferences in the fields.

**Peng Liu** completed his doctoral studies in Economics at the University of Bonn, Germany, in 2019. He is an expert in the emerging field of Computational Economics, with specialized research interests in Data Mining and Machine Learning applications for Economic Forecasting and Macro-Finance. Additionally, Dr. Liu explores the interplay between Social Capital and economic variables, integrating advanced computational techniques to yield actionable insights. He is currently engaged in multidisciplinary research that bridges the traditional boundaries of Economics with state-of-the-art computational methodologies.

**Long Chen** received the B.S. degree in information sciences from Peking University, Beijing, China, in 2000, the M.S.E. degree from the Institute of Automation, Chinese Academy of Sciences, Beijing, in 2003, the M.S. degree in computer engineering from the University of Alberta, Edmonton, AB, Canada, in 2005, and the Ph.D. degree in electrical engineering from the University of Texas at San Antonio, San Antonio, TX, USA, in 2010. From 2010 to 2011, he was a Post-Doctoral Fellow at the University of Texas at San Antonio. He is currently an Associate Professor with the Department of Computer and Information Science, University of Macau, Macau, China. His current research interests include computational intelligence, Bayesian methods, and other machine learning techniques and their applications.

**Witold Pedrycz** received the M.Sc., Ph.D., and D.Sci. degrees in automatic control and computer science from the Silesian University of Technology, Gliwice, Poland, in 1977, 1980, and 1984, respectively. He is currently a Professor and the Canada Research Chair in computational intelligence with the Department of Electrical and Computer Engineering, University of Alberta, Edmonton, AB, Canada. He is also with the Systems Research Institute of the Polish Academy of Sciences, Warsaw, Poland. His main research interests include computational intelligence, fuzzy modeling and granular computing, knowledge discovery and data mining, fuzzy control, pattern recognition, knowledge-based neural networks, relational computing, and software engineering. Dr. Pedrycz was elected as a Foreign Member of the Polish Academy of Sciences in 2009. In 2012, he was elected as a fellow of the Royal Society of Canada. He was a recipient of the prestigious Norbert Wiener Award from the IEEE Systems, Man, and Cybernetics Society in 2007, the IEEE Canada Computer Engineering Medal, the Cajastur Prize for Soft Computing from the European Centre for Soft Computing, the Killam Prize, and the Fuzzy Pioneer Award from the IEEE Computational Intelligence Society. He is also the Editor-in-Chief of Information Sciences, WIREs Data Mining and Knowledge Discovery,and Granular Computing.

**Weiping Ding** received the Ph.D. degree in Computer Science, Nanjing University of Aeronautics and Astronautics, Nanjing, China, in 2013. In 2016, He was a Visiting Scholar at National University of Singapore, Singapore. From 2017 to 2018, he was a Visiting Professor at University of Technology Sydney, Australia. He is a Full Professor with the School of Information Science and Technology, Nantong University, Nantong, China, and also the supervisor of Ph.D postgraduate by the Faculty of Data Science at City University of Macau, China. His main research directions involve deep neural networks, multimodal machine learning, and medical images analysis. He has published over 200 articles, including over 90 IEEE Transactions papers. His fifteen authored/co-authored papers have been selected as ESI Highly Cited Papers. He serves as an Associate Editor/Editorial Board member of IEEE Transactions on Neural Networks and Learning Systems, IEEE Transactions on Fuzzy Systems, IEEE/CAA Journal of Automatica Sinica, IEEE Transactions on Intelligent Transportation Systems, IEEE Transactions on Intelligent Vehicles, IEEE Transactions on Emerging Topics in Computational Intelligence, IEEE Transactions on Artificial Intelligence, Information Fusion, Information Sciences, Neurocomputing, Applied Soft Computing. He is the Leading Guest Editor of Special Issues in several prestigious journals, including IEEE Transactions on Evolutionary Computation, IEEE Transactions on Fuzzy Systems, and Information Fusion